\documentclass[conference]{IEEEtran}
\IEEEoverridecommandlockouts
\usepackage{cite}
\usepackage{amsmath,amssymb,amsfonts}
\usepackage{algorithmic}
\usepackage{graphicx}
\usepackage{comment}
\usepackage{textcomp}
\usepackage{xcolor}
\usepackage{mathtools, nccmath}
\usepackage{color, colortbl}
\usepackage{makecell}
\setcellgapes{1pt}
\renewcommand{\baselinestretch}{0.87}
\DeclarePairedDelimiter{\nint}\lfloor\rceil
\def\BibTeX{{\rm B\kern-.05em{\sc i\kern-.025em b}\kern-.08em
    T\kern-.1667em\lower.7ex\hbox{E}\kern-.125emX}}

\definecolor{LightCyan}{rgb}{0.88,1,1}
\definecolor{Gray}{gray}{0.9}

\begin{document}

\title{Training Energy-Efficient Deep Spiking Neural Networks with Single-Spike Hybrid Input Encoding}


\author{\IEEEauthorblockN{Gourav Datta, Souvik Kundu, Peter A.~Beerel}
\IEEEauthorblockA{\textit{Ming Hsieh Department of Electrical and Computer Engineering} \\
\textit{University of Southern California}\\
Los Angeles, California 90089, USA \\
\{gdatta, souvikku, pabeerel\}@usc.edu}
}

\maketitle

\begin{abstract}

Spiking Neural Networks (SNNs) have emerged as an attractive alternative to traditional deep learning frameworks, since they provide higher computational efficiency in event driven neuromorphic hardware. However, the state-of-the-art (SOTA) SNNs suffer from high inference latency, resulting from inefficient input encoding and training techniques. The most widely used input coding schemes, such as Poisson based rate-coding, do not leverage the temporal learning capabilities of SNNs.  This paper presents a training framework for low-latency energy-efficient SNNs that uses a hybrid encoding scheme at the input layer in which the analog pixel values of an image are directly applied during the first timestep and a novel variant of spike temporal coding is used during subsequent timesteps. In particular, neurons in every hidden layer are restricted to fire at most once per image which increases activation sparsity. To train these hybrid-encoded SNNs, we propose a variant of the gradient descent based spike timing dependent backpropagation (STDB) mechanism using a novel cross entropy loss function based on both the output neurons' spike time and membrane potential. 
The resulting SNNs have reduced latency and high activation sparsity, yielding significant improvements in computational efficiency. In particular, we evaluate our proposed training scheme on image classification tasks from CIFAR-10 and CIFAR-100 datasets on several VGG architectures. We achieve top-1 accuracy of $66.46$\% with $5$ timesteps on the CIFAR-100 dataset with ${\sim}125\times$ less compute energy than an equivalent standard ANN. Additionally, our proposed SNN performs $5$-$300\times$ faster inference compared to other state-of-the-art rate or temporally coded SNN models.

\end{abstract}
\begin{IEEEkeywords}

SNN, STDB, Input encoding, Energy-efficient SNNs
\end{IEEEkeywords}

\section{Introduction}

\label{intro}
Artificial Neural Networks (ANNs) have contributed to a number of impressive success stories in Artificial General Intelligence (AGI) \cite{lecun_dl_nature, dl_microsoft, hinton_nips, self_driving, dl_skin}. However, their superior performance has come at the cost of high computational and memory requirements \cite{nips_45nm, han2015deep}. While convolutional neural networks (CNNs) on general purpose high-performance compute platforms such as GPUs are now ubiquitous \cite{gpu_dl}, there has been increasing interest in domain-specific hardware accelerators \cite{eyeriss_v1} and alternate types of neural networks. In particular, Spiking Neural Network (SNN) accelerators have emerged as a potential low power alternative for AGI \cite{neuro_frontiers,spike_ratecoding,dsnn_conversion1,abhronil_pra}. SNNs attempt to emulate the remarkable energy-efficiency of the brain 
with event-driven neuromorphic hardware. Neurons in an SNN exchange information via discrete binary events, resulting in a significant paradigm shift from traditional CNNs.

Because SNNs receive and transmit information through spikes, analog values must be encoded into a sequence of spikes. There has been a plethora of encoding methods proposed, including rate coding \cite{diehl2016conversion,dsnn_conversion_abhronilfin}, temporal coding \cite{comsa_2020, zhou_2020, park_2020,zhang2020rectified}, rank-order coding \cite{Kheradpisheh_2020}, phase coding \cite{kim_2018}, \cite{park_2019} and other exotic coding schemes \cite{ammar_2019}. Among these, rate-coding has shown competitive performance on complex tasks \cite{diehl2016conversion,dsnn_conversion_abhronilfin} while others are either generally limited to simple tasks such as learning the XOR function and classifying digits from the MNIST dataset or require a large number of spikes for inference. In rate coding, the analog value is converted to a spike train using a Poisson generator function with a rate proportional to the input pixel value. 
The number of timesteps in each train is inversely proportional to the quantization error in the representation, as illustrated in Fig. \ref{fig:lif_model_in_image}(b). Low error requirements force a large number of timesteps at the expense of high inference latency and low activation sparsity \cite{dsnn_conversion_abhronilfin}. Temporal coding, on the other hand, has higher sparsity and can more explicitly represent correlations in inputs. 
However, temporal coding is challenging to scale \cite{Kheradpisheh_2020} to vision tasks and often requires kernel-based spike response models \cite{zhou_2020} which are computationally expensive compared to the traditional leaky-integrate-and-fire (LIF) or integrate-and-fire (IF) models. Recently, the authors in \cite{rathi2020dietsnn} proposed direct input encoding, where they feed the analog pixel values directly into the first convolutional layer, which treats them as input currents to LIF neurons. Another recently proposed temporal encoding scheme uses the discrete cosine transform (DCT) to distribute the spatial pixel information over time for learning low-latency SNNs \cite{garg_2020}. However, up to now, there has been no attempt to combine both spatial (captured by rate or direct encoding) and temporal information processed by the SNNs.  

In addition to accommodating the various of forms of encoding inputs, supervised learning algorithms for SNNs have overcome many roadblocks associated with the discontinuous derivative of the spike activation function \cite{lee_dsnn,wu2019direct,kim_2020}. However, effective SNN training remains a challenge, as seen by the fact that SNNs still lag behind ANNs in terms of latency and accuracy in traditional classification tasks \cite{tavanaei_2019,dsnn_conversion_abhronilfin}. 
A single feed-forward pass in ANN corresponds to multiple forward passes in SNN which is associated with a fixed number of timesteps. In spike-based backpropagation, the backward pass requires the gradients to be integrated over every timestep which increases computation and memory complexity \cite{lee_dsnn, rathi2020iclr}. It requires multiple iterations, is memory intensive (for backward pass computations), and energy-inefficient, and thus has been mainly limited to small datasets (e.g. CIFAR-10) on simple shallow convolutional architectures \cite{rathi2020iclr}. Researchers have also observed high spiking activity and energy consumption in these trained SNN models \cite{kundu_2021}, which further hinders their deployment in edge applications. 
Thus, the current challenges in SNN models are high inference latency and spiking activity, long training time, and high training costs in terms of memory and computation. 

To address these challenges, this paper makes the following contributions:

\begin{itemize}
    \item \textit{Hybrid Spatio-Temporal Encoding:} We employ a hybrid input encoding technique where the real-valued image pixels are fed to the SNN during the first timestep. During the subsequent timesteps, the SNN follows a single-spike temporal coding scheme, where the arrival time of the input spike is inversely proportional to the pixel intensity. While the direct encoding in the first timestep helps the SNN achieve low inference latency, the temporal encoding increases activation sparsity.
    \item \textit{Single Spike LIF Model:} To further harness the benefits of temporal coding, we propose a modified LIF model, where neurons in every hidden layer fire at most once over all the timesteps. This leads to higher activation sparsity and compute efficiency. 
    \item \textit{Novel Loss Function:} We also propose a variant of the gradient descent based spike timing dependent backpropagation mechanism to train SNNs with our proposed encoding technique. In particular, we employ a hybrid cross entropy loss function to capture both the accumulated membrane potential and the spike time of the output neurons.
\end{itemize}

The remainder of our paper is structured as follows. In Section \ref{sec:background} we present the necessary background. Section \ref{sec:hybrid_encoding} describes our proposed input encoding technique. We present our detailed experimental evaluation of the classification accuracy and latency in Section \ref{sec:experiments}. We show the energy improvement of our proposed framework in Section \ref{sec:energy} and finally present conclusions in Section \ref{sec:conc}.

\section{Background}\label{sec:background}

\subsection{SNN Fundamentals}

\begin{figure}[t!]
\begin{center}
\includegraphics[width=0.5\textwidth]{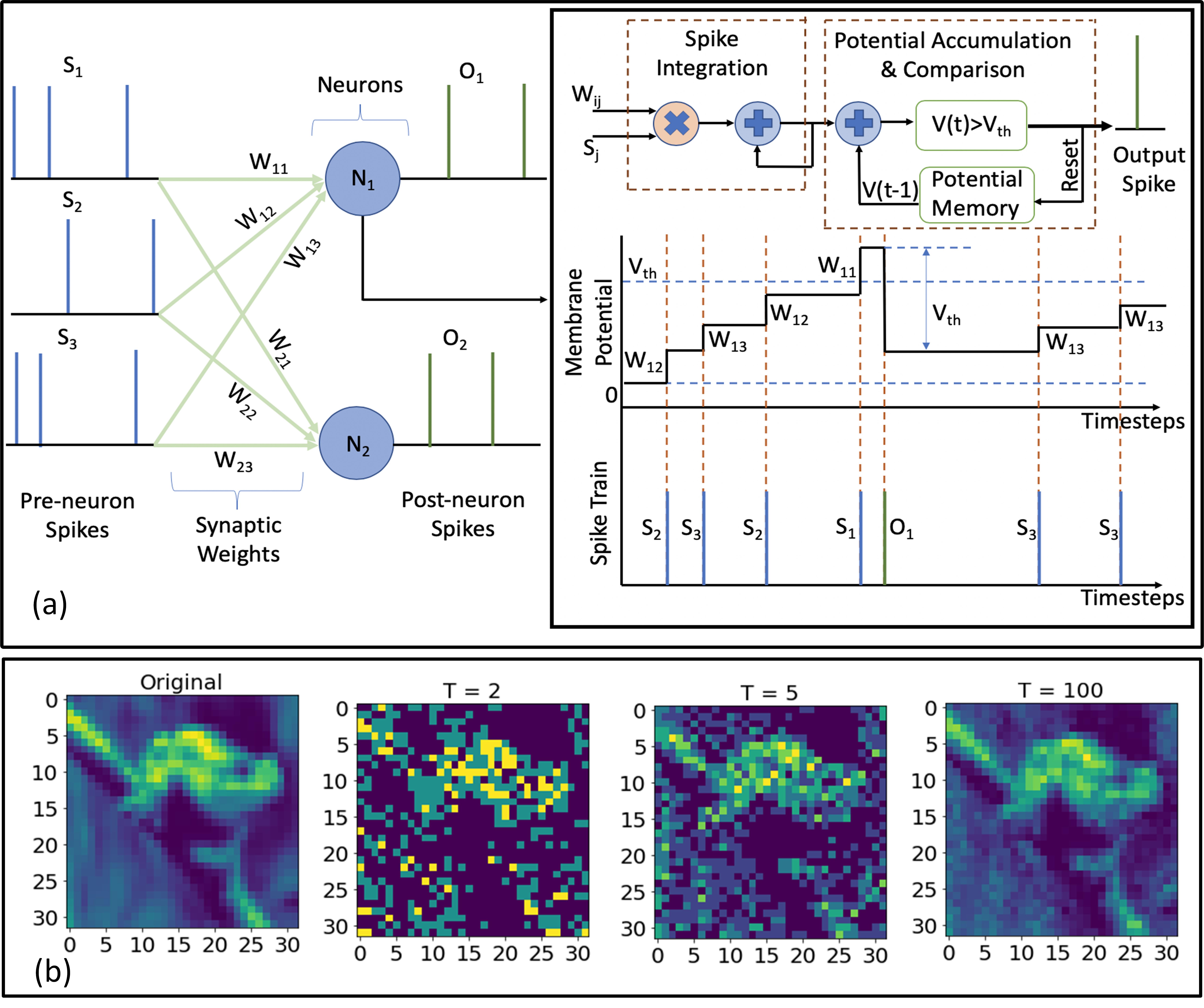}
\end{center}
\caption{(a) Feedforward fully-connected SNN architecture with Integrate and Fire (IF) spiking dynamics, (b) The spike input generated over several timesteps through Poisson generator. It is clear that having more timesteps yields a better approximation of the input image.}
\label{fig:lif_model_in_image}
\vspace{-0.4cm}
\end{figure}

An SNN consists of a network of neurons that communicate through a sequence of spikes modulated by synaptic weights. 
The spiking dynamics of a neuron are typically represented using either Integrate-and-Fire (IF) \cite{lu2020exploring} or Leaky-Integrate-and-Fire (LIF) model \cite{leefin2020}. Fig. \ref{fig:lif_model_in_image}(a) illustrates a basic SNN architecture with IF neurons processing rate-coded inputs. Both IF and LIF neurons integrate the input current into their respective states referred to as membrane potentials. The key difference between the models is that the membrane potential of a IF neuron does not change during the time period between successive input spikes while the LIF neuronal membrane potential leaks with a finite time constant. In this work, we use the LIF model to convert ANNs trained with ReLU activations to SNNs, because the leaky behaviour 
provides improved robustness to noisy spike-inputs and better generalization compared to those with no leak \cite{chowdhury_2020}. Moreover, the leak term provides a tunable control knob, which can be leveraged to improve inference accuracy, latency, and spiking activity in SNNs.

To characterize the LIF model, we use the following differential equation 

\begin{equation}
C\frac{dU_i^t}{dt}+GU_i^t=I_i^t=\sum_{j} W_{ij}\cdot{S_j^t}
\label{eq:IF_continuous}
\end{equation}
\noexpand
where $C$ and $G$ are the membrane capacitance and conductance respectively. $U_i^t$ and $I_i^t$ are the membrane potential and input synaptic current of the $i^{th}$ neuron at time $t$. Note that $U_i^t$ integrates the incoming (pre-neuron) spikes $S_j^t$ modulated by weights $W_{ij}$ and leaks with a time constant equal to $\frac{C}{G}$. The post-neuron generates an output spike when $U_i$ exceeds the firing threshold $V$. However, because of its' continuous representation, Eq. \ref{eq:IF_continuous} is not suitable for implementations in popular Machine Learning (ML) frameworks (eg. Pytorch). Hence, we convert Eq. \ref{eq:IF_continuous} into an iterative discrete-time version, as shown in Eq. \ref{eq:LIF_discrete} \cite{rathi2020iclr}, in which spikes are characterized as binary values (1 represents the presence of a spike). Note that $\lambda$ represents the leak term which reduces $U_i$ by a factor of $(1-\lambda)$ in every timestep.

\begin{equation}
U_i^t=\lambda U_i^{t-1}+\sum_j W_{ij}{S_j(t)}-V{O_i^{t-1}}
\label{eq:LIF_discrete}
\end{equation}
The binary output spike at timestep $t$ is given as 
\begin{equation}
O_i^t=
\begin{cases}
    1, & \text{if } U_i^t>V\\
    0, & \text{otherwise}
\end{cases}
\label{eq:IF_out_spike}
\end{equation}
Note that the last term in Eq. \ref{eq:LIF_discrete} represents soft reset that reduces the membrane potential $U_i$ by the threshold $V$ at timestep $t$ in response to an output spike generated at timestep $(t-1)$. In contrast, hard reset means resetting $U_i$ to $0$ after an output spike is generated. Soft reset minimises the information loss by allowing the spiking neuron to carry forward the surplus potential above the firing threshold to the subsequent timestep \cite{rathi2020iclr, ledinauskas2020training} and is adopted in this work.

\subsection{SNN Training Techniques}

Recent research on training supervised deep SNNs can be broadly divided into three categories: i) Indirect learning; ii) Direct Learning; iii) Hybrid Learning.

\subsubsection{Indirect Learning}

Recent works have demonstrated that SNNs can be efficiently converted from ANNs by approximating the activation value of ReLU neurons with the firing rate of spiking neurons \cite{dsnn_conversion1, dsnn_conversion_cont,dsnn_conversion_ijcnn,dsnn_conversion_abhronilfin,dsnn_conversion5}. This technique uses the standard backpropagation algorithm for training in the ANN domain, and helps SNNs achieve SOTA results on various challenging inference tasks, particularly in image recognition \cite{dsnn_conversion_cont,dsnn_conversion_abhronilfin}. Moreover, ANN-SNN conversion simplifies the training procedures compared to approximate gradient techniques, since it involves only a single forward pass to process a single input. However, a disadvantage of ANN-SNN conversion is that it yields SNNs with an order of magnitude higher latency than other training techniques \cite{dsnn_conversion_abhronilfin}. 
In this work, we use ANN-SNN conversion as an initial step in our proposed framework because it yields high classification accuracy on deep networks. We then leverage direct encoding in the first timestep to reduce the number of synaptic operations and thus improve the SNN's energy efficiency.

\subsubsection{Direct Learning}

The discontinuous and non-differentiable nature of a spiking neuron makes it difficult to implement gradient descent based backpropagation. Consequently, several approximate training methodologies have been proposed that leverage the temporal dynamics of SNNs 
\cite{connor_sensf, lee_dsnn, lee_stdp, panda2016_sup, bellec_2018long, neftci_surg}. The basic idea of these works is to approximate the spiking neuron functionality with a continuous differentiable model or use surrogate gradients to approximate  real gradients. 
However, STDB requires the gradients to be integrated over all timesteps, increasing computation and memory requirements significantly, particularly for deep networks.

\subsubsection{Hybrid Learning}

Authors in \cite{rathi2020iclr} proposed a hybrid training methodology that consists of ANN-SNN conversion, followed by approximate gradient descent on the initialized network to obtain the final trained SNN model. The authors claimed that combining the two training techniques  helps SNNs converge within a few epochs and require fewer timesteps.
Another recent paper \cite{rathi2020dietsnn} proposes a training scheme for deep SNNs in which the membrane leak and the firing threshold along with other network parameters (weights) are updated at the end of every batch via gradient descent after ANN-SNN conversion. Moreover, instead of converting the image pixel values into spike trains using Poisson rate coding described above, the authors directly feed the analog pixel values in the first convolutional layer, which emits spikes using the LIF neuron model. This enables requiring fewer timesteps compared to Poisson rate coding. In this work, we employ a variant of the hybrid learning technique (ANN-SNN Conversion, followed by STDB with trainable weights, threshold and leak) to train deep SNNs.


\section{Hybrid Spike Encoding}\label{sec:hybrid_encoding}

\begin{figure}[t!]
\begin{center}
\includegraphics[width=0.47\textwidth]{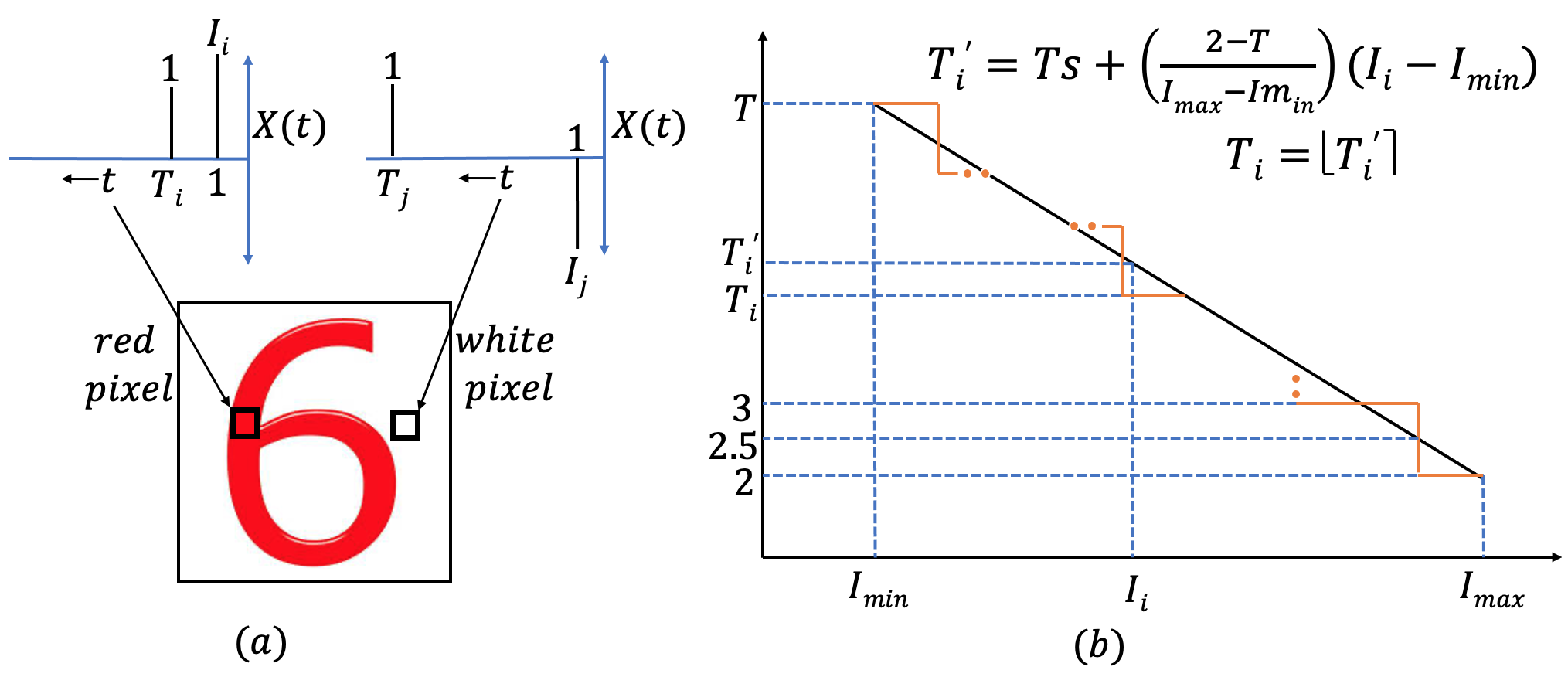}
\end{center}
\caption{(a) Hybrid coded input to the SNN (b) Mapping between the pixel intensity of images and the firing time of individual neurons where $\nint{.}$ denotes the nearest integer function}

\label{fig:hybrid_encoding}
\vspace{-0.4cm}
\end{figure}

We propose a hybrid encoding scheme to convert the real-valued pixel intensities of input images into SNN inputs over the total number of timesteps dictated by the desired inference accuracy. As is typical, input images fed to the ANN are normalized to zero mean and unit standard deviation. 
In our proposed coding technique, we feed the analog pixel value in the input layer of the SNN in the $1^{st}$ timestep. Next, we convert the real-valued pixels into a spike train starting from the $2^{nd}$ timestep representing the same information. Considering a gray image with pixel intensity values in the range $[I_{min},I_{max}]$, each input neuron encodes the temporal information of its' corresponding pixel value in a single spike time in the range $[2,T]$ where $T$ is the total number of timesteps. The firing time of the $i^{th}$ input neuron, $T_i$, is computed based on the $i^{th}$ pixel intensity value, $I_i$, as follows

\begin{equation}\label{eq-hyrbid_spikecoding}
T_i=\nint{T+\left(\frac{2-T}{I_{max}-I_{min}}\right)\cdot(I_i-I_{min})}
\end{equation}
\noexpand
where $\nint{.}$ represents the nearest integer function. Eq. \ref{eq-hyrbid_spikecoding} is represented as the point-slope form of the linear relationship shown in Fig. \ref{fig:hybrid_encoding}(b) and $\nint{.}$ is applied because $T_i$ should be integral. Note that Eq. \ref{eq-hyrbid_spikecoding} also implies that the spike train starts from the $2^{nd}$ timestep $2$. The encoded value of the $i^{th}$ neuron in the input layer is thus expressed as

\begin{align}
    X_i(t)=
    \begin{cases}
      I_i,& \text{if } t=1 \\
      1,& \text{else if } t=T_i \\
      0,& \text{otherwise}
      \end{cases}
      \end{align}
      \noexpand
which is further illustrated in Fig. \ref{fig:hybrid_encoding}(b). Brighter image pixels have higher intensities, and hence, lower $T_i$. Neurons at the subsequent layers fire as soon as they reach their threshold, and both the voltage potential and time to reach the threshold in the output layer determines the network decision. The analog pixel value in the $1^{st}$ time step influences the membrane potential of the output neurons, while the firing times of the input neurons based on the pixel intensities are responsible for the spike times of the output neurons.

Notably, this hybrid encoding scheme captures both the intensity and temporal nature of the input neurons, does not need any preprocessing steps like applying Gabor filters that are commonly used in SNNs trained with spike time dependent plasticity (STDP)
\cite{Kheradpisheh_2018, mozafari_2018}. Moreover, our proposed encoding technique is compatible with event-driven cameras which capture actual pixel value first, and subsequently emit spikes based on the changes in pixel intensity \cite{gallego_2020}. Lastly, our proposal ensures that there is a single input spike per pixel, and hence, the obtained spike train is sparser than that observed in rate/direct coded techniques.


\section{Proposed Training Scheme}\label{sec:proposed_training}

We employ a modified version of the LIF model illustrated in Section \ref{sec:background} to train energy-efficient SNNs. In our proposed training framework, neurons in all the hidden convolutional and fully-connected layers (except the output layer) spike at most once over all the timesteps. During inference, once a neuron emits a spike, it is shut off, and does not participate in the remaining LIF computations. However, during training, the neurons in the hidden layers follow the model illustrated in Eq. (6)-(8) which shows that even though each neuron can fire at most once, it still needs to perform computations following the LIF model. This ensures that the error gradients are still non-zero following the spike time and enables our proposed training framework to avoid the dead neuron problem where learning does not happen in the absence of a spike.

\begin{align}
    \mbox{\boldmath $U$}_l^t&=\lambda_l{\mbox{\boldmath $U$}_l^{t-1}}+W_l{\mbox{\boldmath $O$}_{l-1}^t}-V_l\cdot(\mbox{\boldmath $z$}_l^{t-1}>0)\label{eq:lif_1} \\
    \mbox{\boldmath $z$}_l^t&=\frac{\mbox{\boldmath $U$}_l^t}{V_l}-1  \label{eq:lif_2}
    \end{align}
    \begin{equation}\label{eq:lif_3}
           \mbox{\boldmath $O$}_l^t =
    \begin{cases}
    1, & \text{if } \mbox{\boldmath $z$}_l^t>0 \text{ and } \mbox{\boldmath $z$}_l^{t_i} \leq 0 \ {\forall} t_i \in [1,t) \\
    0, & \text{otherwise }
       \end{cases}
    \end{equation}
    \noexpand
\noexpand    
Note that $\mbox{\boldmath $U$}_l^t$, $\mbox{\boldmath $O$}_{{l}-{1}}$, and $W_l$ are vectors containing the membrane potential of the neurons of layer $l$ at timestep $t$, spike signals from layer $({l}-{1})$, and the weight matrix connecting the layer $l$ and $({l}-{1})$. Also note that $(\mbox{\boldmath $z$}_l^{t-1}>0)$ in Eq. \ref{eq:lif_1} denotes a Boolean vector of size equal to the number of neurons in layer $l$. The leak and threshold voltage for all the neurons in layer $l$ are represented by $\lambda_l$ and $V_l$ respectively. In our training framework, both these parameters (same for the neurons in a particular layer) are trained with backpropagation along with the weights to optimize both accuracy and latency. 

 The neurons in the output layer accumulate the incoming inputs without any leakage as shown in Eq. \ref{eq:lif_output}. However, unlike previous works \cite{rathi2020dietsnn,rathi2020iclr}, the output neurons in our proposed framework emit spikes following a model shown in Eq. \ref{eq:t_last_layer} where $\mbox{\boldmath $T$}_l$ denote the vector containing the spike times of the output neurons and $T$ is the total number of timesteps.

\begin{align}
    \mbox{\boldmath $U$}_l^t &=\mbox{\boldmath $U$}_l^{t-1}+W_l\mbox{\boldmath $O$}_{l-1}^t \label{eq:lif_output} \\
    \mbox{\boldmath $T$}_l &= 
    \begin{cases}
        T, & \text{if }   \mbox{\boldmath $U$}_l^T<V_l \\
        \mbox{\boldmath $t$} \ s.t. \ \mbox{\boldmath $U$}_l^t \ {\geq} \ { V_l} \ \& \ \mbox{\boldmath $U$}_l^{t-1}<{V_l}, & \text{otherwise} \label{eq:t_last_layer}
    \end{cases}
\end{align}
\noexpand
The output layer only triggers an output spike if there was no spike in the earlier timesteps and the corresponding membrane potential crosses the threshold. Also, an output neuron is forced to fire at the last timestep if it was unable to emit a spike in any of the timesteps. This ensures that all the neurons in the output layer have a valid $T_l$ which can be included in the loss function. 

Let us now we derive the expressions to compute the gradients of the trainable parameters of all the layers. We perform the spatial and temporal credit assignment by unrolling the network in temporal axis and employing backpropagation through time (BPTT) \cite{rathi2020iclr}.  \\
\textit{Output Layer:} The loss function is defined on both $\mbox{\boldmath $U$}_l^T$ and $\mbox{\boldmath $T$}_l$ to correctly capture both the direct and temporal information presented at the input layer. Therefore, we employ two softmax functions of the $i^{th}$ output neuron shown in Eq. \ref{eq:softmax}, where $N$ denotes the total number of classes, $U_i^T$ and $t_i$ represent the accumulated membrane potential after the final timestep and the firing time of the $i^{th}$ neuron respectively. . 
\begin{equation}\label{eq:softmax}
    \tilde{U_i}=\frac{e^{U_i^T}}{\sum_{j=1}^{N}e^{U_j^T}}, \   \ \tilde{t_i}={\frac{e^{-t_i}}{\sum_{j=1}^{N}e^{-t_j}}}
\end{equation}


The resulting hybrid cross entropy loss ($\mathcal{L}$) and its' gradient with respect to the accumulated membrane potential vector ($\frac{\partial\mathcal{L}}{\partial\mbox{\boldmath $U$}_l^T}$) are thus defined as

\begin{equation}
    \mathcal{L}=-\sum_{i=1}^N{y_ilog(\tilde{U_i}\tilde{t_i})}, \quad  \frac{\partial\mathcal{L}}{\partial\mbox{\boldmath $U$}_l^T}=\mbox{\boldmath $\tilde{U}$}_l^T-\mbox{\boldmath $y$}
\end{equation}
\noexpand
where $\mbox{\boldmath $\tilde{U}$}_l^T$ is the vector containing the softmax values $\tilde{U}_i$, and $\mbox{\boldmath $y$}$ is the one-hot encoded vector of the correct class. Similarly, the gradient with respect to the firing time vector ($\frac{\partial\mathcal{L}}{\partial\mbox{\boldmath $T$}_l}$) is $(\mbox{\boldmath $\tilde{T}$}_l-\mbox{\boldmath $y$})$. Now, we compute the weight update as

\begin{align}
W_l&=W_l-\eta\Delta{W_l}\\
\Delta{W_l}&=\sum_{t}\frac{\partial\mathcal{L}}{\partial W_l}=\sum_{t}\frac{\partial\mathcal{L}}{\partial\mbox{\boldmath $U$}_l^t}\frac{\partial\mbox{\boldmath $U$}_l^t}{\partial W_l} \notag \\
&=\frac{\partial\mathcal{L}}{\partial\mbox{\boldmath $U$}_l^T}\sum_{t}\frac{\partial\mbox{\boldmath $U$}_l^t}{\partial W_l}=(\mbox{\boldmath $\tilde{U}$}_l^T-\mbox{\boldmath $y$})\sum_{t}\mbox{\boldmath $O$}_{l-1}^t
\end{align}
\noexpand
where $\eta$ is the learning rate (LR). In order to evaluate the threshold update at the output layer, we rewrite Eq. (10) as 

\begin{align}
    \mbox{\boldmath $T$}_l=\sum_{t=1}^{T-1}(t\mathcal{H}(a)\mathcal{H}(b))+T\mathcal{H}(c)
\end{align}
\noexpand
where $\mathcal{H}$ denotes the Heaviside step function, $\mbox{\boldmath $a$}=\mbox{\boldmath $U$}_l^t-\mbox{\boldmath $V$}_l$, $\mbox{\boldmath $b$}=\mbox{\boldmath $V$}_l-\mbox{\boldmath $U$}_l^{t-1}$, and $\mbox{\boldmath $c$}=\mbox{\boldmath $V$}_l-\mbox{\boldmath $U$}_l^T$. Note that $\mbox{\boldmath $V$}_l$ represents a vector of repeated elements of the threshold voltage of the output layer. The derivative $(\frac{\partial\mbox{\boldmath $T$}_l}{\partial V_l})$ can then be represented as
\begin{align}
    \frac{\partial\mbox{\boldmath $T$}_l}{\partial V_l}=\sum_{t=1}^{T-1}t(\mathcal{H}(\mbox{\boldmath $a$})\delta(\mbox{\boldmath $b$})-\mathcal{H}(\mbox{\boldmath $b$})\delta(\mbox{\boldmath $a$}))+ T\delta(\mbox{\boldmath $c$})
\end{align}
where $\delta$ represents the Dirac-delta function. Since the delta function is zero almost everywhere, it will not allow the gradient of $\mbox{\boldmath $T$}_l$ to change and train $V_l$. Hence, we approximate Eq. (16) as 
\begin{align}
    \sum_{t=1}^{T-1}{t(\mathcal{H}(\mbox{\boldmath $a$})(|\mbox{\boldmath $b$}|{<}\mbox{\boldmath $\beta$}){-}\mathcal{H}(\mbox{\boldmath $b$})(|\mbox{\boldmath $a$}|{<}\mbox{\boldmath $\beta$}]){+}T(|\mbox{\boldmath $c$}|{<}\mbox{\boldmath $\beta$})}
\end{align}
\noexpand
where $\mbox{\boldmath $\beta$}$ is a vector of size equal to the number of output neurons, consisting of the repeated elements of a training hyperparameter that controls the gradient of $\mbox{\boldmath $T$}_l$. Note that $(|\mbox{\boldmath $a$}|{<}\mbox{\boldmath $\beta$})$,$(|\mbox{\boldmath $b$}|{<}\mbox{\boldmath $\beta$})$, and $(|\mbox{\boldmath $c$}|{<}\mbox{\boldmath $\beta$}$) are all Boolean vectors of the same size as $\mbox{\boldmath $\beta$}$. We then compute the threshold update as 
\begin{align}
V_l=V_l-\eta\Delta{V_l},  \ \Delta{V_l}=\frac{\partial\mathcal{L}}{\partial V_l}=\frac{\partial\mathcal{L}}{\partial\mbox{\boldmath $T$}_l}\frac{\partial\mbox{\boldmath $T$}_l}{\partial V_l}
\end{align}

\textit{Hidden Layers:} 
The weight update of the $l^{th}$ hidden layer is calculated from Eq. (6)-(8) as

\begin{align}
    \Delta{W_l}&=\sum_{t}\frac{\partial\mathcal{L}}{\partial W_l}=\sum_{t}\frac{\partial\mathcal{L}}{\partial\mbox{\boldmath $z$}_l^t}\frac{\partial\mbox{\boldmath $z$}_l^t}{\partial\mbox{\boldmath $O$}_l^t}\frac{\partial\mbox{\boldmath $O$}_l^t}{\partial\mbox{\boldmath $U$}_l^t}\frac{\partial\mbox{\boldmath $U$}_l^t}{\partial W_l} \notag \\
    &=\sum_{t}\frac{\partial\mathcal{L}}{\partial\mbox{\boldmath $z$}_l^t}\frac{\partial\mbox{\boldmath $z$}_l^t}{\partial\mbox{\boldmath $O$}_l^t}\frac{1}{V_l}\mbox{\boldmath $O$}_{l-1}^t
\end{align}
\noexpand
$\frac{d\mbox{\boldmath $z$}_l^t}{d\mbox{\boldmath $O$}_l^t}$ is the non-differentiable gradient which can be approximated with the surrogate gradient proposed in \cite{bellec_2018long}.

\begin{align}
    \frac{\partial\mbox{\boldmath $z$}_l^t}{\partial\mbox{\boldmath $O$}_l^t}=\gamma\cdot{max(0,1-|\mbox{\boldmath $z$}_l^t|)}
\end{align}

\noindent
where $\gamma$ is a hyperparameter denoting the maximum value of the gradient. The threshold update is then computed as

\begin{align}
   \Delta{V_l}&=\sum_{t}\frac{\partial\mathcal{L}}{\partial V_l}=\sum_{t}\frac{\partial\mathcal{L}}{\partial\mbox{\boldmath $O$}_l^t}\frac{\partial\mbox{\boldmath $O$}_l^t}{\partial\mbox{\boldmath $z$}_l^t}\frac{\partial\mbox{\boldmath $z$}_l^t}{\partial V_l} \notag \\ &=\sum_{t}\frac{\partial\mathcal{L}}{\partial\mbox{\boldmath $O$}_l^t}\frac{\partial\mbox{\boldmath $O$}_l^t}{\partial\mbox{\boldmath $z$}_l^t}\left(\frac{-V_l\cdot(\mbox{\boldmath $z$}_l^{t-1}>0)-\mbox{\boldmath $U$}_l^t}{(V_l)^2}\right)
\end{align}

Given that the threshold is same for all neurons in a particular layer, it may seem redundant to train both the weights and threshold together. However, our experimental evaluation detailed in Section \ref{sec:energy} shows that the number of timesteps required to obtain the state-of-the-art classification accuracy decreases with this joint optimization. We hypothesize that this is because the optimizer is able to reach an improved local minimum when both parameters are tunable. Finally, the leak update is computed as

\begin{align}
   \lambda_l&=\lambda_l-\eta\Delta{\lambda_l} \\
   \Delta\lambda_l=\sum_{t}\frac{\partial\mathcal{L}}{\partial\lambda_l}&=\sum_{t}\frac{\partial\mathcal{L}}{\partial \mbox{\boldmath $O$}_l^t}\frac{\partial \mbox{\boldmath $O$}_l^t}{\partial \mbox{\boldmath $z$}_l^t}\frac{\partial \mbox{\boldmath $z$}_l^t}{\partial \mbox{\boldmath $U$}_l^t}\frac{\partial \mbox{\boldmath $U$}_l^t}{\partial \lambda_l} \notag\\
   &=\sum_{t}\frac{\partial\mathcal{L}}{\partial \mbox{\boldmath $O$}_l^t}\frac{\partial \mbox{\boldmath $O$}_l^t}{\partial \mbox{\boldmath $z$}_l^t}\frac{1}{V_l}\mbox{\boldmath $U$}_l^{(t-1)}
\end{align}

\section{Experiments}\label{sec:experiments}
\label{sec:expt}

This section first describes how we evaluate the efficacy of our proposed encoding and training framework and then presents the inference accuracy on CIFAR-10 and CIFAR-100 datasets with various VGG model variants.

\subsection{Experimental Setup}
\label{subsec:setup}


\subsubsection{ANN Training for Initialization}

To train our ANNs, we used the standard data-augmented input set for each model. 
For ANN training with various VGG models, we imposed a number of constraints that leads to near lossless SNN conversion \cite{dsnn_conversion_abhronilfin}. In particular, our models are trained without the bias term because it complicates parameter space exploration which increases conversion difficulty and tends to increase conversion loss. The absence of bias term implies that Batch Normalization \cite{bjorck2018understanding} cannot be used as a regularizer during the training process. 
Instead, we use Dropout \cite{srivastava_2014} as the regularizer for both ANN and SNN training. Also, our pooling operations use average pooling because for binary spike based activation layers, max pooling incurs significant information loss. 
We performed the ANN training for $200$ epochs with an initial LR of $0.01$ that decays by a factor of $0.1$ after $120$, $160$, and $180$ epochs.   

\subsubsection{ANN-SNN Conversion and SNN Training}

Previous works \cite{dsnn_conversion_abhronilfin,rathi2020iclr} set the layer threshold of the first hidden layer by computing the maximum input to a neuron over all its neurons across all $T$ timesteps for a set of input images \cite{dsnn_conversion_abhronilfin}. The thresholds of the subsequent layers are sequentially computed in a similar manner taking the maximum across all neurons and timesteps. However, in our proposed framework, the threshold for each layer is computed sequentially as the $99.7$ percentile (instead of the maximum) of the neuron input distribution at each layer, which improves the SNN classification accuracy \cite{rathi2020dietsnn}. 
During threshold computation, the leak in the hidden layers is set to unity and the analog pixel values of an image are directly applied to the input layer \cite{rathi2020dietsnn}. We considered only $512$ input images to limit conversion time and used a threshold scaling factor of $0.4$ for SNN training and inference, following the recommendations in \cite{rathi2020iclr}.

Initialized with these layer thresholds and the trained ANN weights, we performed our proposed SNN training with the hybrid input encoding scheme for $150$ epochs for CIFAR-10 and CIFAR-100, respectively, where we jointly optimize the weights, the membrane leak, and the firing thresholds of each layer as described in Section \ref{sec:proposed_training}. We set $\gamma$ = $0.3$ \cite{bellec_2018long}, $\beta=0.2$, and used a starting LR of $10^{-4}$ which decays by a factor of $0.1$ every $10$ epochs.

\begin{table}[!t]
\caption{Model performances with single-spike hybrid encoded SNN training on CIFAR-10 and CIFAR-100 after a) ANN training, b) ANN-to-SNN conversion and c) SNN training.}
\begin{center}
\scriptsize\addtolength{\tabcolsep}{-3.5pt}
\begin{tabular}{|c|c|c|c|c|c|}
\hline
 & a. & {b. Accuracy ($\%$) with} & c. Accuracy ($\%$) after \\
 Architecture & ANN ($\%$) & {ANN-SNN conversion} &  proposed SNN training\\
 & accuracy     &  for $T$ = 200  &    for  T=5   \\ 
\hline
\hline
 \multicolumn{4}{|c|}{Dataset : CIFAR-10} \\
\hline
\hline
 VGG-6   & 90.22 & 89.98 & 88.89 \\
\hline
 VGG-11   & 91.02  & 91.77  &  90.66  \\
\hline
 VGG-16  & 93.24 & 93.16  & 91.41 \\
\hline
\hline
 \multicolumn{4}{|c|}{Dataset : CIFAR-100} \\
\hline
\hline
 VGG-16  & 71.02 & 70.38 & 66.46  \\

\hline
\end{tabular}
\end{center}
\label{tab:snn_acc_results}
\vspace{-0.4cm}
\end{table}

\subsection{Classification Accuracy \& Latency}

We evaluated the performance of these networks on multiple VGG architectures, namely VGG-6, VGG-9 and VGG-11 for CIFAR-10 and VGG-16 for CIFAR-100 datasets respectively. Column-$2$ in Table \ref{tab:snn_acc_results} shows the ANN accuracy; column-$3$ shows the accuracy after ANN-SNN conversion with $200$ timesteps. Note that we need $200$ timesteps to evaluate the thresholds of the SNN for VGG architectures without any significant loss in accuracy. Column-4 in Table \ref{tab:snn_acc_results} shows the accuracy when we perform our proposed training with our hybrid input encoding discussed in Section \ref{sec:hybrid_encoding}. The performance of the SNNs trained via our proposed framework is compared with the current state-of-the-art SNNs with various encoding and training techniques in Table \ref{tab:snn_comparison}. Our proposal requires only $5$ timesteps for both SNN training and inference to obtain the SOTA test accuracy and hence, representing $5$-$300 \times$ improvement in inference latency compared to other rate/temporally coded spiking networks. Note that the direct encoding in the first time step is crucial for SNN convergence, and temporal coding solely leads to a test accuracy of ${\sim}10\%$ and ${\sim}1\%$ on CIFAR-10 and CIFAR-100 respectively, for all the network architectures. 

\begin{table}
\caption{Performance comparison of the proposed single spike hybrid encoded SNN with state-of-the-art deep SNNs on CIFAR-10 and CIFAR-100. TTFS denotes time-to-first-spike coding.}
\begin{center}
\scriptsize\addtolength{\tabcolsep}{-1.5pt}
\begin{tabular}{|c|c|c|c|c|c|}
\hline
Authors &  Training & Input & Architecture & Accuracy & Time  \\
 & type & encoding  &  & ($\%$) & steps \\
\hline
\hline
 \multicolumn{6}{|c|}{Dataset : CIFAR-10} \\
\hline
\hline
Sengupta et  & ANN-SNN & Rate  &VGG-16 & 91.55 & 2500 \\
al. (2019) \cite{dsnn_conversion_abhronilfin} & conversion & &  & & \\
\hline
Wu et al. & Surrogate & Direct & 5 CONV,  & 90.53 & 12 \\
(2019) \cite{wu2019direct} & gradient & & 2 linear &  & \\
\hline
Rathi et al. & Conversion+ & Rate & VGG-16  & 91.13 & 100 \\
(2020) \cite{rathi2020iclr} & STDB training  & & & 92.02 & 200 \\
\hline
Garg et al. & Conversion+ & DCT & VGG-9 & 89.94 & 48 \\
(2019) \cite{garg_2020} &STDB  training & & & & \\
\hline
Kim et al. & ANN-SNN & Phase& VGG-16 & 91.2 & 1500 \\
(2018) \cite{kim_2018}& conversion & & &  & \\
\hline
Park et al. & ANN-SNN & Burst & VGG-16 & 91.4 & 1125 \\
(2019) \cite{park_2019}& conversion & & &  & \\
\hline
Park et al. & STDB & TTFS & VGG-16 & 91.4 & 680 \\
(2020) \cite{park_2020}& training & & &  & \\
\hline
Kim at. al. & Surrogate & Rate & VGG-9 & 90.5 & 25 \\
(2020) \cite{kim_2020}& gradient & & & &  \\
\hline
Rathi at. al. & Conversion+ & Direct & VGG-16 & 92.70 & 5 \\
(2020) \cite{rathi2020dietsnn}& STDB training & & & \textbf{93.10} & 10 \\
\hline
\rowcolor{Gray}
This work & Conversion+ &Hybrid & VGG-16 &  {91.41} & {5} \\
\rowcolor{Gray}
& STDB training & & & & \\
\hline
\hline
 \multicolumn{5}{|c|}{Dataset : CIFAR-100} \\
\hline
\hline
Lu et al. & ANN-SNN & Direct& VGG-16 & 63.20 & 62 \\
(2020) \cite{lu2020exploring}& conversion & & &  & \\
\hline
Garg et al. & Conversion+ & DCT & VGG-11 & 68.3 & 48 \\
(2020) \cite{garg_2020} &STDB  training & & & & \\
\hline
Park et al. & ANN-SNN & Burst & VGG-16 & 68.77 & 3100 \\
(2019) \cite{park_2019}& conversion & & &  & \\
\hline
Park et al. & STDB & TTFS & VGG-16 & 68.8 & 680 \\
(2020) \cite{park_2020}& training & & &  & \\
\hline
Kim at. al. & Surrogate & Rate & VGG-9 & 66.6 & 50 \\
(2020) \cite{kim_2020}& gradient & & & &  \\
\hline
Rathi et al. & Conversion+ & Direct & VGG-16 & \textbf{69.67} & 5 \\
(2020) \cite{rathi2020dietsnn}& STDB training  & & &  & \\
\hline
\rowcolor{Gray}
This work & Conversion+ & Hybrid & VGG-16 & {66.46} & {5} \\
\rowcolor{Gray}
& STDB training & & & & \\
\hline
\end{tabular}
\end{center}
\label{tab:snn_comparison}
\vspace{-0.6cm}
\end{table}

\section{Improvement in Energy-efficiency}\label{sec:energy}
\label{subsec:energy}

\subsection{Reduction in Spiking Activity}

To model energy consumption, we assume a generated SNN spike consumes a fixed amount of energy \cite{dsnn_conversion1}. Based on this assumption, earlier works \cite{rathi2020iclr, dsnn_conversion_abhronilfin} have adopted the average spiking activity (also known as average spike count) of an SNN layer $l$, denoted ${\zeta}^l$, as a measure of compute-energy of the model. In particular, ${\zeta}^l$ is computed as the ratio of the total spike count in $T$ steps over all the neurons of the layer $l$ to the total number of neurons in that layer. Thus lower the spiking activity, the better the energy efficiency.

\begin{figure}[t!]
\begin{center}
\includegraphics[width=0.36\textwidth]{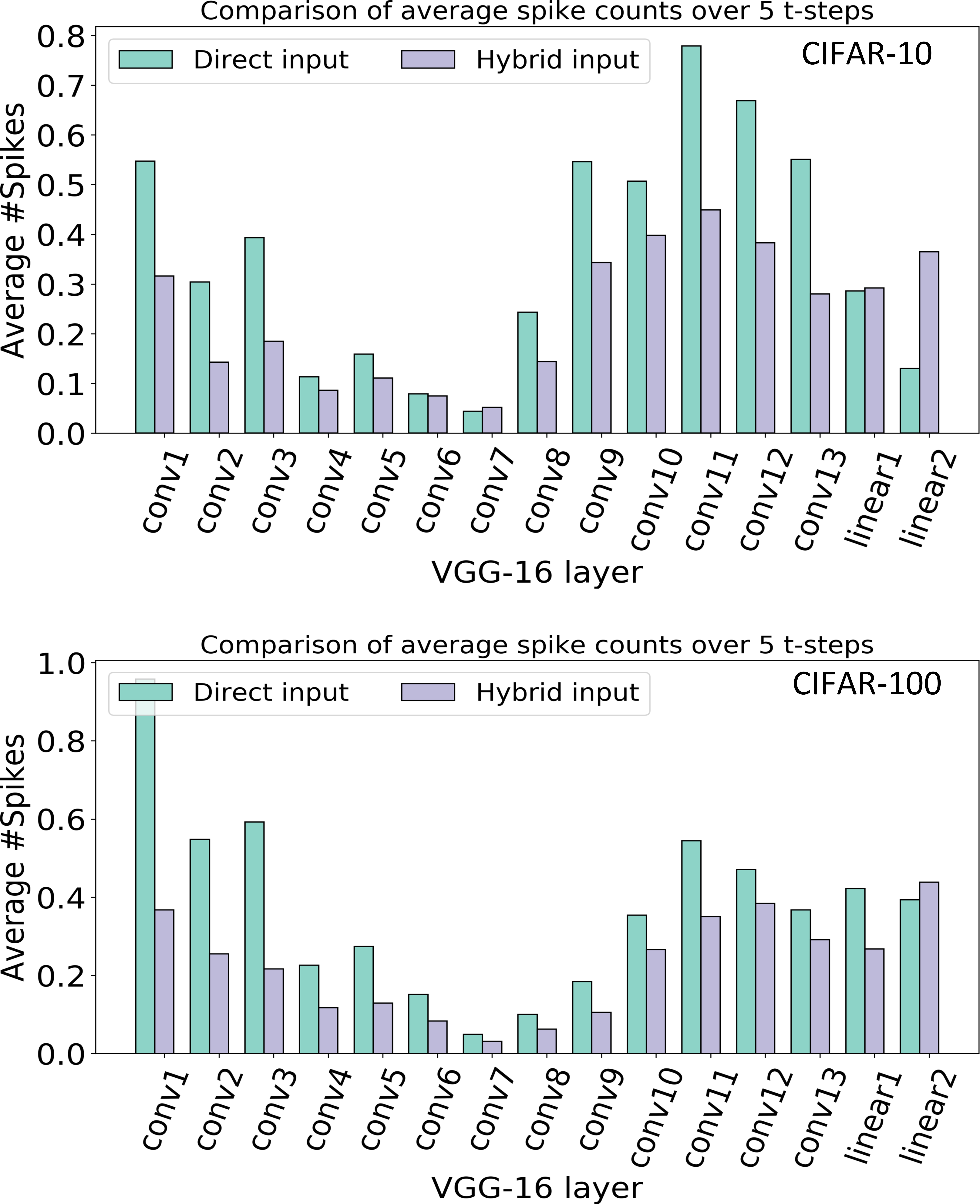}
\end{center}
\vspace{-0.2cm}
\caption{Comparison of average spiking activity per layer for VGG-16 on CIFAR-10 and CIFAR-100 with both direct and hybrid input encoding.}
\vspace{-0.2cm}
\label{fig:hybrid_vs_drect_spike}
\end{figure}

Fig. \ref{fig:hybrid_vs_drect_spike} shows the average number of spikes for each layer with our proposed single-spike hybrid encoding and direct encoding scheme on VGG-16 when evaluated for 1500 samples from CIFAR-10 testset for VGG-16 architecture. Let the average be denoted by $\zeta^l$ which is computed by summing all the spikes in a layer over 100 timesteps and dividing by the number of neurons in that layer. For example, the average spike count of the $11^{th}$ convolutional layer of the direct encoded SNN is $0.78$, which implies that over a $5$ timestep period each neuron in that layer spikes $0.78$ times on average over all input samples. 
As we can see, the spiking activity for almost all the layers reduces significantly with our proposed encoding technique. 

\begin{figure}[t!]
\begin{center}
\includegraphics[width=0.40\textwidth]{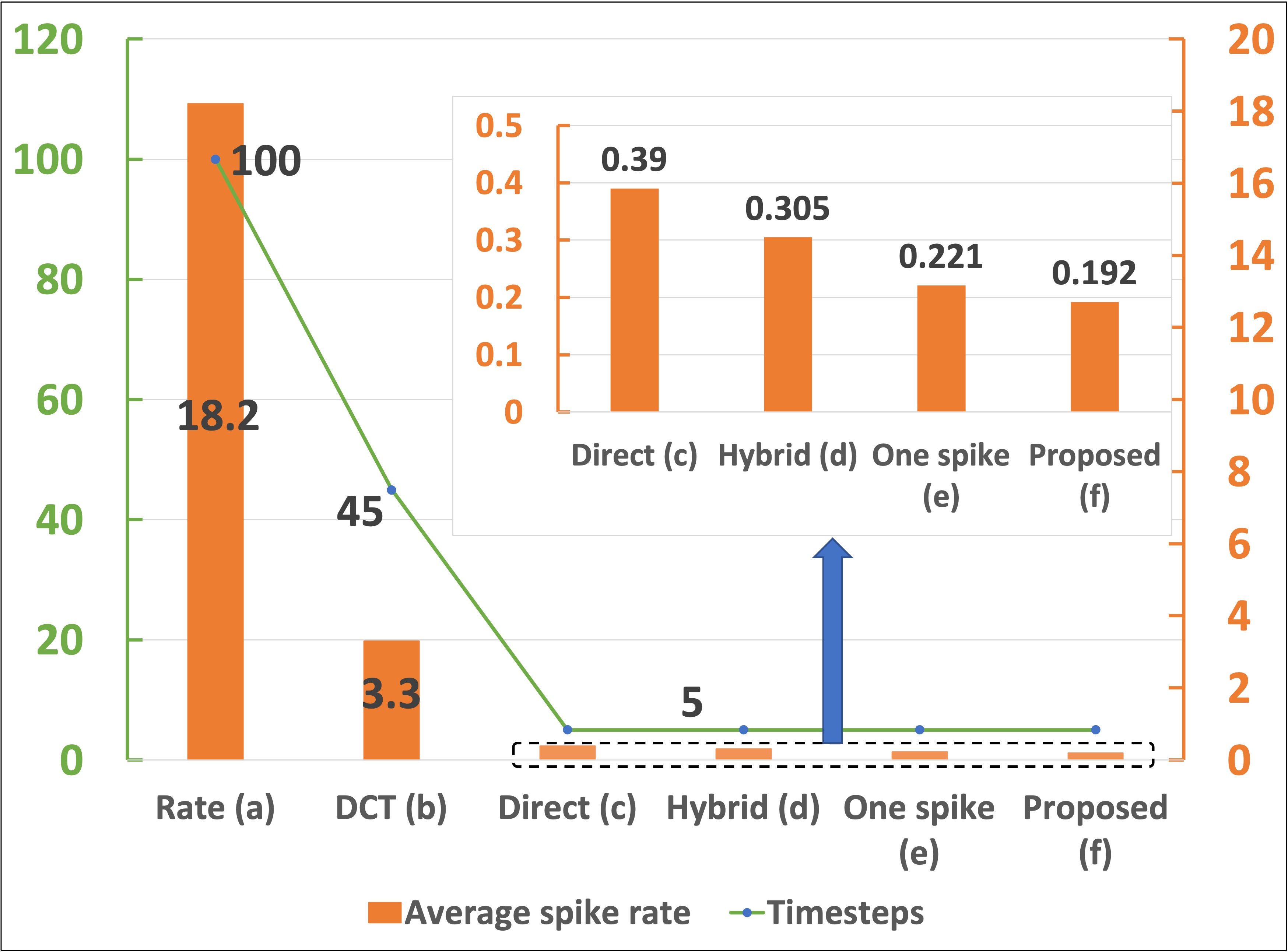}
\end{center}
\vspace{-0.2cm}
\caption{Effect of Poisson rate encoding, DCT encoding, direct encoding, and single-spike hybrid input encoding on the average spike rate and latency for VGG-16 architecture on CIFAR-10 dataset.  }
\label{fig:ablation_prev_work}
\vspace{-0.3cm}
\end{figure}

To compare our proposed work with the SOTA SNNs, we perform hybrid training (ANN-SNN conversion, along with STDB) on spiking networks with (a) IF neurons with Poisson rate encoding \cite{rathi2020iclr}, (b) IF neurons with DCT-based input encoding \cite{garg_2020}, and (c) LIF neurons with direct encoding \cite{rathi2020dietsnn}. We employ trainable leak and threshold in these SNNs for fair comparison. We also evaluate the impact of the hybrid spatio-temporal encoding with the modified loss function and the single-spike constraint individually on the average spike rate and latency under similar accuracy and conditions (trainable threshold and leak). In particular, we train three additional spiking networks: (d) SNN with LIF neuron and proposed hybrid-encoding, (e) SNN with LIF neuron and direct encoding with the single-spike constraint over all the layers, and (f) single-spike hybrid encoded SNN with LIF neuron. All the six networks achieve test accuracies between $90${-}$93\%$ for VGG-16 on CIFAR-10. Fig. \ref{fig:ablation_prev_work} shows the average spiking rate and the number of timesteps required to obtain the SOTA test accuracy of all these SNNs. Both (d) and (e) result in lower average spiking activity compared to all the SOTA SNNs, with at most the same number of timesteps. Finally, (f) generates even lower number of average spikes ($2\times$,$17.2\times$, and $94.8\times$ less compared to direct, DCT, and rate coding)  with the lowest inference latency reported till date for deep SNN architectures \cite{rathi2020dietsnn}, and no significant reduction in the test accuracy. The improvement stems from both the hybrid input encoding which reduces spiking activity in the initial few layers and our single-spike constraint which reduces the average spike rate throughout the network, particularly in the later layers. It becomes increasingly difficult for the membrane potential of the convolutional layers deep into the network to increase sufficient to emit a spike, due to the fact that the neurons in the earlier layers cannot fire multiple times and we need only $5$ timesteps for classification.

\begin{table}[!ht]
\caption{Convolutional and Fully-connected layer FLOPs for ANN and SNN models}
\scriptsize\addtolength{\tabcolsep}{-4pt}
\begin{center}
\begin{tabular}{|c|c|c|c|}
\hline
{Model} & \multicolumn{3}{|c|}{Number of FLOPs}\\
\cline{2-4}
        & Notation & Convolutional layer $l$ & Fully-connected layer $l$ \\
\hline
{$ANN$} &  $F_{ANN}^l$ & $(k^l)^2\times H_o^l\times W_o^l\times C_o^l\times C_i^l$ & $f_i^l\times f_o^l$\\
\hline
{$SNN$} & $F_{SNN}^l$ & $(k^l)^2\times H_o^l\times W_o^l\times C_o^l\times C_i^l \times {\zeta}^l$ & $f_i^l\times f_o^l \times {\zeta}^l$  \\
\hline
\end{tabular}
\label{tab:flops}

\end{center}
\vspace{-0.4cm}
\end{table}

\subsection{Reduction in FLOPs and Compute Energy}

Let us assume a convolutional layer $l$ having weight tensor ${\textbf{W}^l}\in{\mathbb{R}^{k^l\times k^l\times C_{i}^l\times C_{o}^l}}$ that operates on an input activation tensor $\textbf{I}^l \in \mathbb{R}^{H_{i}^l \times W_{i}^l \times C_{i}^l}$, where $ H_{i}^l, W_{i}^l$, $C_i^l$ and $C_{o}^l$ are the input tensor height, width, number of channels, and filters, respectively. $k^l$ represents both filter height and width. We now quantify the energy consumed to produce the corresponding output activation tensor $\textbf{O}^l \in \mathbb{R}^{H_{o}^l \times W_{o}^l \times C_{o}^l}$ for an ANN and SNN, respectively. Our model can be extended to fully-connected layers with $f_{i}^{l}$ and $f_{o}^{l}$ as the number of input and output features respectively.
In particular, for an ANN, the total number of FLOPS for layer $l$, denoted $F_{ANN}^l$, is shown in row 1 of Table \ref{tab:flops} \cite{kundu2020pre, kundu2019psconv}. The formula can be easily adjusted for an SNN in which the number of FLOPs at layer $l$ is a function of the average spiking activity at the layer $(\zeta^l)$ denoted as $F_{SNN}^l$ in Table \ref{tab:flops}. Thus, as the activation output gets sparser, the compute energy decreases.  

For ANNs, FLOPs primary consist of multiply accumulate (MAC) operations of the convolutional and linear layers. On the contrary, for SNNs, except the first and last layer, the FLOPs are limited to accumulates (ACs) as the spikes are binary and thus simply indicate which weights need to be accumulated at the post-synaptic neurons. For the first layer, we need to use MAC units as we consume analog 
input\footnote{Note that for the hybrid coded data input we need to perform MAC at the first layer at $t=1$, and AC operation during remaining timesteps at that layer. For the direct coded input, only MAC during the $1^{st}$ timestep is sufficient, as neither the inputs nor the weights change during remaining timesteps (i.e. $ 5 \geq t \geq 2 $).} (at timestep one). Hence, the compute energy for an ANN $(E_{ANN})$ and an iso-architecture SNN model $(E_{SNN})$ can be written as 
\begin{align}
     E_{ANN} &=(\sum_{l=1}^{L}F^l_{SNN})\cdot{E_{MAC}}\\
    E_{SNN} &=(F^1_{ANN})\cdot{E_{MAC}}+(\sum_{l=2}^{L}F^l_{SNN})\cdot{E_{AC}}
\end{align}
\noindent
where $L$ is the total number of layers. Note that $E_{MAC}$ and $E_{AC}$ are the energy consumption for a MAC and AC operation respectively. As shown in Table \ref{tab:fp_energy}, $E_{AC}$ is $\mathord{\sim}32 \times$ lower than $E_{MAC}$ \cite{horowitz20141} in $45$ nm CMOS technology. This number may vary for different technologies, but generally, in most technologies, an AC operation is significantly cheaper than a MAC operation. 

\begin{figure}[t!]
\begin{center}
\includegraphics[width=0.48\textwidth]{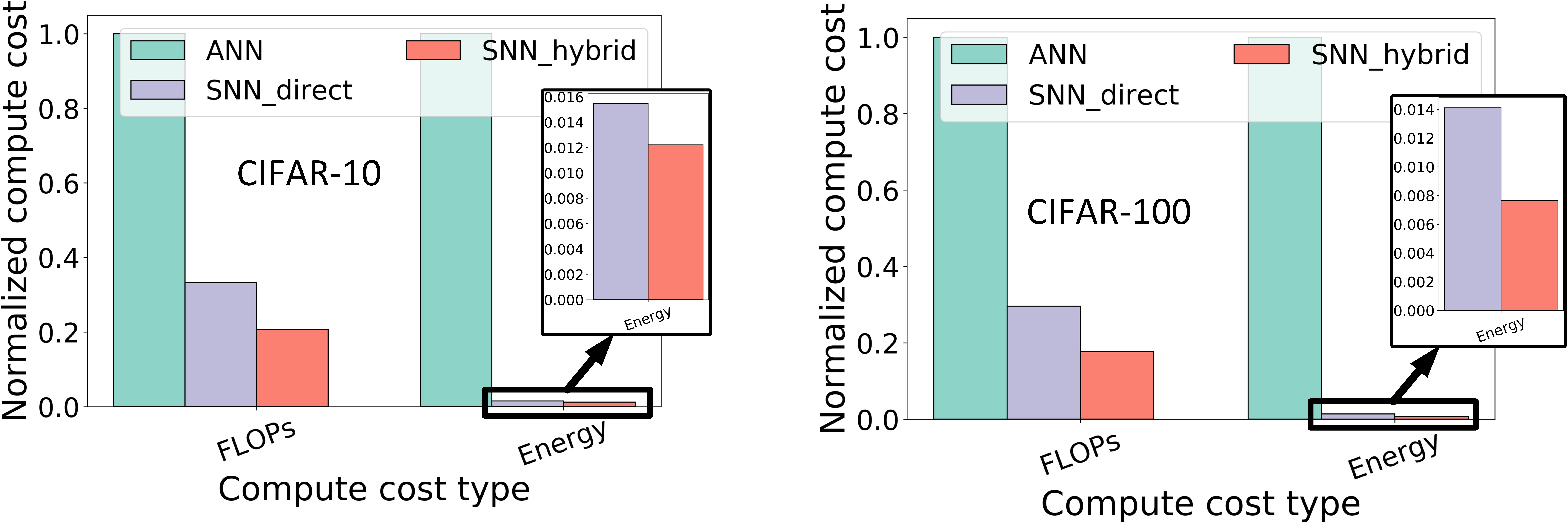}
\end{center}
\vspace{-0.1cm}
\caption{Comparison of normalized compute cost on CIFAR-10 and CIFAR-100 for VGG-16 of ANN and SNN with direct and hybrid input encoding.}
\vspace{-0.2cm}
\label{fig:hybrid_vs_drect_compute_cost}
\end{figure}

Fig. \ref{fig:hybrid_vs_drect_compute_cost} illustrates the energy consumption and FLOPs for ANN and SNN models of VGG-16 while classifying the CIFAR datasets, where the energy is normalized to that of an equivalent ANN.
The number of FLOPs for SNNs obtained by our proposed training framework is smaller than that for an ANN with similar number of parameters. Moreover, because the ACs consume significantly less energy than MACs, as shown in Table \ref{tab:fp_energy}, SNNs are significantly more energy efficient. In particular, for CIFAR-10 our proposed SNN consumes $\mathord{\sim}70 \times$ less compute energy than a comparable iso-architecture ANN with similar parameters and $\mathord{\sim}1.2 \times$ less compute energy than a comparable SNN with direct encoding technique and trainable threshold/leak \cite{rathi2020dietsnn} parameters. For CIFAR-100 with hybrid encoding and our single-spike constraint, the energy-efficiency can reach up to $\mathord{\sim}125 \times$ and $\mathord{\sim}1.8 \times$, respectively, compared to ANN and direct-coded SNN models \cite{rathi2020dietsnn} having similar parameters and architecture. Note that we did not consider the memory access energy in our evaluation because it is dependent on the underlying system architecture. Although SNNs incur significant data movement because the membrane potentials need to be fetched at every timestep, there have been many proposals to reduce the memory cost by data buffering \cite{shen_2017}, computing in non-volatile crossbar memory arrays \cite{chen2015crossbar}, and data reuse with energy-efficient dataflows \cite{eyeriss_v1}. All these techniques can be applied to the SNNs obtained by our proposed training framework to address the memory cost.

\section{Conclusions}\label{sec:conc}

SNNs that operate with discrete spiking events can potentially unlock the energy wall in deep learning for edge applications. Towards this end, we presented a training framework that leads to low latency, energy-efficient spiking networks with high activation sparsity. We initialize the parameters of our proposed SNN taken from a trained ANN, to speed-up the training with spike-based backpropagation. The image pixels are applied directly as input to the network during the first timestep, while they are converted to a sparse spike train with firing times proportional to the pixel intensities in subsequent timesteps. We also employ a modified version of the LIF model for the hidden and output layers of the SNN, in which all the neurons fire at most once per image. Both of these lead to high activation sparsity in the input, convolutional, and dense layers of the network. Moreover, we employ a hybrid cross entropy loss function to account for the spatio-temporal encoding in the input layer and train the network weights, firing threshold, and membrane leak via spike-based backpropagation to optimize both accuracy and latency. The high sparsity combined with low inference latency reduces the compute energy by ${\sim}70$-$130\times$ and ${\sim}1.2$-$1.8\times$ compared to an equivalent ANN and a direct encoded SNN respectively with similar accuracy. SNNs obtained by our proposed framework achieves similar accuracy as other state-of-the-art rate or temporally coded SNN models with $5$-$300\times$ fewer timesteps. Future works include power and performance evaluation of our energy-efficient models on neurmorphic chips such as Loihi \cite{loihi2018} and exploration of neuromorphic datasets \cite{neuromorphicdataset2019} to leverage the temporal learning ability of our training framework. 

\begin{table}
\caption{Estimated energy costs for MAC and AC operations in 45 $nm$ CMOS process at 0.9 V \cite{horowitz20141}}
\scriptsize\addtolength{\tabcolsep}{-4pt}
\begin{center}
\begin{tabular}{|c|c|c|}
\hline
{Serial No.} & {Operation} & {Energy ($pJ$)} \\
\hline
1. & 32-bit multiplication $int$ & $3.1$ \\
2. & 32-bit addition $int$ & $0.1$ \\
\hline
3. & 32-bit MAC &  $3.2$ ($\#1$ + $\#2$) \\
4. & 32-bit AC  & $0.1$ ($\#2$)\\
\hline
\end{tabular}
\label{tab:fp_energy}
\vspace{-0.8cm}
\end{center}
\end{table}

\section{Acknowledgements}

This work was supported in part by the NSF CCF-1763747 award.

\bibliographystyle{IEEEtran}
\bibliography{IEEEabrv,biblio}

\begin{thebibliography}{10}
\providecommand{\url}[1]{#1}
\csname url@samestyle\endcsname
\providecommand{\newblock}{\relax}
\providecommand{\bibinfo}[2]{#2}
\providecommand{\BIBentrySTDinterwordspacing}{\spaceskip=0pt\relax}
\providecommand{\BIBentryALTinterwordstretchfactor}{4}
\providecommand{\BIBentryALTinterwordspacing}{\spaceskip=\fontdimen2\font plus
\BIBentryALTinterwordstretchfactor\fontdimen3\font minus
  \fontdimen4\font\relax}
\providecommand{\BIBforeignlanguage}[2]{{%
\expandafter\ifx\csname l@#1\endcsname\relax
\typeout{** WARNING: IEEEtran.bst: No hyphenation pattern has been}%
\typeout{** loaded for the language `#1'. Using the pattern for}%
\typeout{** the default language instead.}%
\else
\language=\csname l@#1\endcsname
\fi
#2}}
\providecommand{\BIBdecl}{\relax}
\BIBdecl

\bibitem{lecun_dl_nature}
Y.~LeCun, Y.~Bengio, and G.~Hinton, ``Deep learning,'' \emph{Nature}, vol. 521,
  pp. 436--44, 05 2015.

\bibitem{dl_microsoft}
l.~Deng, J.~Li, J.-T. Huang, K.~Yao, D.~Yu, F.~Seide, M.~Seltzer, G.~Zweig,
  X.~He, J.~Williams, Y.~Gong, and A.~Acero, ``Recent advances in deep learning
  for speech research at microsoft,'' in \emph{Acoustics, Speech, and Signal
  Processing, 1988. ICASSP-88., 1988 International Conference on}, 10 2013, pp.
  8604--8608.

\bibitem{hinton_nips}
A.~Krizhevsky, I.~Sutskever, and G.~E. Hinton, ``Imagenet classification with
  deep convolutional neural networks,'' in \emph{Advances in Neural Information
  Processing Systems 25}, F.~Pereira, C.~J.~C. Burges, L.~Bottou, and K.~Q.
  Weinberger, Eds.\hskip 1em plus 0.5em minus 0.4em\relax Curran Associates,
  Inc., 2012, pp. 1097--1105.

\bibitem{self_driving}
C.~{Chen}, A.~{Seff}, A.~{Kornhauser}, and J.~{Xiao}, ``{DeepDriving}: Learning
  affordance for direct perception in autonomous driving,'' in \emph{2015 IEEE
  International Conference on Computer Vision (ICCV)}, vol.~1, no.~1, 2015, pp.
  2722--2730.

\bibitem{dl_skin}
A.~Rezvantalab, H.~Safigholi, and S.~Karimijeshni, ``Dermatologist level
  dermoscopy skin cancer classification using different deep learning
  convolutional neural networks algorithms,'' \emph{arXiv preprint
  arXiv:1810.10348}, 2018.

\bibitem{nips_45nm}
S.~Han, J.~Pool, J.~Tran, and W.~J. Dally, ``Learning both weights and
  connections for efficient neural networks,'' in \emph{Proceedings of the 28th
  International Conference on Neural Information Processing Systems - Volume
  1}.\hskip 1em plus 0.5em minus 0.4em\relax MIT Press, 2015, p. 1135–1143.

\bibitem{han2015deep}
S.~Han, H.~Mao, and W.~J. Dally, ``Deep compression: Compressing deep neural
  networks with pruning, trained quantization and {Huffman} coding,''
  \emph{arXiv preprint arXiv:1510.00149}, 2015.

\bibitem{gpu_dl}
S.~Lym, D.~Lee, M.~O’Connor, N.~Chatterjee, and M.~Erez, ``{DeLTA}: {GPU}
  performance model for deep learning applications with in-depth memory system
  traffic analysis,'' \emph{2019 IEEE International Symposium on Performance
  Analysis of Systems and Software (ISPASS)}, Mar 2019.

\bibitem{eyeriss_v1}
Y.-H. Chen, J.~Emer, and V.~Sze, ``{Eyeriss}: A spatial architecture for
  energy-efficient dataflow for convolutional neural networks,'' in \emph{ACM
  SIGARCH Computer Architecture News}, vol.~44, 06 2016.

\bibitem{neuro_frontiers}
G.~Indiveri and T.~Horiuchi, ``Frontiers in neuromorphic engineering,''
  \emph{Frontiers in Neuroscience}, vol.~5, 2011.

\bibitem{spike_ratecoding}
M.~Pfeiffer and T.~Pfeil, ``Deep learning with spiking neurons: Opportunities
  and challenges,'' \emph{Frontiers in Neuroscience}, vol.~12, p. 774, 2018.

\bibitem{dsnn_conversion1}
Y.~Cao, Y.~Chen, and D.~Khosla, ``Spiking deep convolutional neural networks
  for energy-efficient object recognition,'' \emph{International Journal of
  Computer Vision}, vol. 113, pp. 54--66, 05 2015.

\bibitem{abhronil_pra}
A.~Sengupta, A.~Banerjee, and K.~Roy, ``Hybrid spintronic-{CMOS} spiking neural
  network with on-chip learning: Devices, circuits, and systems,'' \emph{Phys.
  Rev. Applied}, vol.~6, Dec 2016.

\bibitem{diehl2016conversion}
P.~U. Diehl, G.~Zarrella, A.~Cassidy, B.~U. Pedroni, and E.~Neftci,
  ``Conversion of artificial recurrent neural networks to spiking neural
  networks for low-power neuromorphic hardware,'' in \emph{2016 IEEE
  International Conference on Rebooting Computing (ICRC)}.\hskip 1em plus 0.5em
  minus 0.4em\relax IEEE, 2016, pp. 1--8.

\bibitem{dsnn_conversion_abhronilfin}
A.~Sengupta, Y.~Ye, R.~Wang, C.~Liu, and K.~Roy, ``Going deeper in spiking
  neural networks: {VGG} and residual architectures,'' \emph{Frontiers in
  Neuroscience}, vol.~13, p.~95, 2019.

\bibitem{comsa_2020}
I.~M. {Comsa}, K.~{Potempa}, L.~{Versari}, T.~{Fischbacher}, A.~{Gesmundo}, and
  J.~{Alakuijala}, ``Temporal coding in spiking neural networks with alpha
  synaptic function,'' in \emph{ICASSP 2020 - 2020 IEEE International
  Conference on Acoustics, Speech and Signal Processing (ICASSP)}, vol.~1,
  no.~1, 2020, pp. 8529--8533.

\bibitem{zhou_2020}
S.~Zhou, X.~LI, Y.~Chen, S.~T. Chandrasekaran, and A.~Sanyal, ``Temporal-coded
  deep spiking neural network with easy training and robust performance,''
  \emph{arXiv preprint arXiv:1909.10837}, 2020.

\bibitem{park_2020}
S.~Park, S.~Kim, B.~Na, and S.~Yoon, ``{T2FSNN}: Deep spiking neural networks
  with time-to-first-spike coding,'' \emph{arXiv preprint arXiv:2003.11741},
  2020.

\bibitem{zhang2020rectified}
M.~Zhang, J.~Wang, B.~Amornpaisannon, Z.~Zhang, V.~Miriyala, A.~Belatreche,
  H.~Qu, J.~Wu, Y.~Chua, T.~E. Carlson, and H.~Li, ``Rectified linear
  postsynaptic potential function for backpropagation in deep spiking neural
  networks,'' 2020.

\bibitem{Kheradpisheh_2020}
S.~R. Kheradpisheh and T.~Masquelier, ``Temporal backpropagation for spiking
  neural networks with one spike per neuron,'' \emph{International Journal of
  Neural Systems}, vol.~30, no.~06, May 2020.

\bibitem{kim_2018}
J.~Kim, H.~Kim, S.~Huh, J.~Lee, and K.~Choi, ``Deep neural networks with
  weighted spikes,'' \emph{Neurocomputing}, vol. 311, pp. 373--386, 2018.

\bibitem{park_2019}
S.~{Park}, S.~{Kim}, H.~{Choe}, and S.~{Yoon}, ``Fast and efficient information
  transmission with burst spikes in deep spiking neural networks,'' in
  \emph{2019 56th ACM/IEEE Design Automation Conference (DAC)}, vol.~1, no.~1,
  2019, pp. 1--6.

\bibitem{ammar_2019}
D.~Almomani, M.~Alauthman, M.~Alweshah, O.~Dorgham, and F.~Albalas, ``A
  comparative study on spiking neural network encoding schema: implemented with
  cloud computing,'' \emph{Cluster Computing}, vol.~22, 06 2019.

\bibitem{rathi2020dietsnn}
N.~Rathi and K.~Roy, ``{DIET-SNN}: Direct input encoding with leakage and
  threshold optimization in deep spiking neural networks,'' \emph{arXiv
  preprint arXiv:2008.03658}, 2020.

\bibitem{garg_2020}
I.~Garg, S.~S. Chowdhury, and K.~Roy, ``{DCT-SNN}: Using {DCT} to distribute
  spatial information over time for learning low-latency spiking neural
  networks,'' \emph{arXiv preprint arXiv:2010.01795}, 2020.

\bibitem{lee_dsnn}
J.~H. Lee, T.~Delbruck, and M.~Pfeiffer, ``Training deep spiking neural
  networks using backpropagation,'' \emph{Frontiers in Neuroscience}, vol.~10,
  2016.

\bibitem{wu2019direct}
Y.~Wu, L.~Deng, G.~Li, J.~Zhu, Y.~Xie, and L.~Shi, ``Direct training for
  spiking neural networks: Faster, larger, better,'' in \emph{Proceedings of
  the AAAI Conference on Artificial Intelligence}, vol.~33, 2019, pp.
  1311--1318.

\bibitem{kim_2020}
Y.~Kim and P.~Panda, ``Revisiting batch normalization for training low-latency
  deep spiking neural networks from scratch,'' \emph{arXiv preprint
  arXiv:2010.01729}, 2020.

\bibitem{tavanaei_2019}
A.~Tavanaei, M.~Ghodrati, S.~R. Kheradpisheh, T.~Masquelier, and A.~Maida,
  ``Deep learning in spiking neural networks,'' \emph{Neural Networks}, vol.
  111, p. 47–63, Mar 2019.

\bibitem{rathi2020iclr}
N.~Rathi, G.~Srinivasan, P.~Panda, and K.~Roy, ``Enabling deep spiking neural
  networks with hybrid conversion and spike timing dependent backpropagation,''
  \emph{arXiv preprint arXiv:2005.01807}, 2020.

\bibitem{kundu_2021}
S.~Kundu, G.~Datta, M.~Pedram, and P.~A. Beerel, ``Spike-thrift: Towards
  energy-efficient deep spiking neural networks by limiting spiking activity
  via attention-guided compression,'' in \emph{Proceedings of the IEEE/CVF
  Winter Conference on Applications of Computer Vision (WACV)}, January 2021,
  pp. 3953--3962.

\bibitem{lu2020exploring}
S.~Lu and A.~Sengupta, ``Exploring the connection between binary and spiking
  neural networks,'' \emph{arXiv preprint arXiv:2002.10064}, 2020.

\bibitem{leefin2020}
C.~Lee, S.~S. Sarwar, P.~Panda, G.~Srinivasan, and K.~Roy, ``Enabling
  spike-based backpropagation for training deep neural network architectures,''
  \emph{Frontiers in Neuroscience}, vol.~14, p. 119, 2020.

\bibitem{chowdhury_2020}
S.~S. Chowdhury, C.~Lee, and K.~Roy, ``Towards understanding the effect of leak
  in spiking neural networks,'' \emph{arXiv preprint arXiv:2006.08761}, 2020.

\bibitem{ledinauskas2020training}
E.~Ledinauskas, J.~Ruseckas, A.~Juršėnas, and G.~Buračas, ``Training deep
  spiking neural networks,'' \emph{arXiv preprint arXiv:2006.04436}, 2020.

\bibitem{dsnn_conversion_cont}
B.~Rueckauer, I.-A. Lungu, Y.~Hu, M.~Pfeiffer, and S.-C. Liu, ``Conversion of
  continuous-valued deep networks to efficient event-driven networks for image
  classification,'' \emph{Frontiers in Neuroscience}, vol.~11, p. 682, 2017.

\bibitem{dsnn_conversion_ijcnn}
P.~U. {Diehl}, D.~{Neil}, J.~{Binas}, M.~{Cook}, S.~{Liu}, and M.~{Pfeiffer},
  ``Fast-classifying, high-accuracy spiking deep networks through weight and
  threshold balancing,'' in \emph{2015 International Joint Conference on Neural
  Networks (IJCNN)}, vol.~1, no.~1, 2015, pp. 1--8.

\bibitem{dsnn_conversion5}
Y.~Hu, H.~Tang, and G.~Pan, ``Spiking deep residual network,'' \emph{arXiv
  preprint arXiv:1805.01352}, 2018.

\bibitem{connor_sensf}
P.~O'Connor, D.~Neil, S.-C. Liu, T.~Delbruck, and M.~Pfeiffer, ``Real-time
  classification and sensor fusion with a spiking deep belief network,''
  \emph{Frontiers in neuroscience}, vol.~7, p. 178, 2013.

\bibitem{lee_stdp}
C.~Lee, P.~Panda, G.~Srinivasan, and K.~Roy, ``Training deep spiking
  convolutional neural networks with {STDP}-based unsupervised pre-training
  followed by supervised fine-tuning,'' \emph{Frontiers in Neuroscience},
  vol.~12, 2018.

\bibitem{panda2016_sup}
P.~Panda and K.~Roy, ``Unsupervised regenerative learning of hierarchical
  features in spiking deep networks for object recognition,'' \emph{arXiv
  preprint arXiv:1602.01510}, 2016.

\bibitem{bellec_2018long}
G.~Bellec, D.~Salaj, A.~Subramoney, R.~Legenstein, and W.~Maass, ``Long
  short-term memory and learning-to-learn in networks of spiking neurons,''
  \emph{arXiv preprint arXiv:1803.09574}, 2018.

\bibitem{neftci_surg}
E.~O. {Neftci}, H.~{Mostafa}, and F.~{Zenke}, ``Surrogate gradient learning in
  spiking neural networks: Bringing the power of gradient-based optimization to
  spiking neural networks,'' \emph{IEEE Signal Processing Magazine}, vol.~36,
  no.~6, pp. 51--63, 2019.

\bibitem{Kheradpisheh_2018}
\BIBentryALTinterwordspacing
S.~R. Kheradpisheh, M.~Ganjtabesh, S.~J. Thorpe, and T.~Masquelier,
  ``{STDP-based} spiking deep convolutional neural networks for object
  recognition,'' \emph{Neural Networks}, vol.~99, p. 56–67, Mar 2018.
  [Online]. Available: \url{http://dx.doi.org/10.1016/j.neunet.2017.12.005}
\BIBentrySTDinterwordspacing

\bibitem{mozafari_2018}
M.~{Mozafari}, S.~R. {Kheradpisheh}, T.~{Masquelier}, A.~{Nowzari-Dalini}, and
  M.~{Ganjtabesh}, ``First-spike-based visual categorization using
  reward-modulated {STDP},'' \emph{IEEE Transactions on Neural Networks and
  Learning Systems}, vol.~29, no.~12, pp. 6178--6190, 2018.

\bibitem{gallego_2020}
G.~{Gallego}, T.~{Delbruck}, G.~M. {Orchard}, C.~{Bartolozzi}, B.~{Taba},
  A.~{Censi}, S.~{Leutenegger}, A.~{Davison}, J.~{Conradt}, K.~{Daniilidis},
  and D.~{Scaramuzza}, ``Event-based vision: A survey,'' \emph{IEEE
  Transactions on Pattern Analysis and Machine Intelligence}, vol.~1, no.~1,
  pp. 1--1, 2020.

\bibitem{bjorck2018understanding}
N.~Bjorck, C.~P. Gomes, B.~Selman, and K.~Q. Weinberger, ``Understanding batch
  normalization,'' in \emph{Advances in Neural Information Processing Systems},
  2018, pp. 7694--7705.

\bibitem{srivastava_2014}
N.~Srivastava, G.~Hinton, A.~Krizhevsky, I.~Sutskever, and R.~Salakhutdinov,
  ``Dropout: A simple way to prevent neural networks from overfitting,''
  \emph{Journal of Machine Learning Research}, vol.~15, pp. 1929--1958, 06
  2014.

\bibitem{kundu2020pre}
S.~{Kundu}, M.~{Nazemi}, M.~{Pedram}, K.~M. {Chugg}, and P.~A. {Beerel},
  ``Pre-defined sparsity for low-complexity convolutional neural networks,''
  \emph{IEEE Transactions on Computers}, vol.~69, no.~7, pp. 1045--1058, 2020.

\bibitem{kundu2019psconv}
S.~Kundu, S.~Prakash, H.~Akrami, P.~A. Beerel, and K.~M. Chugg, ``{pSConv}: A
  pre-defined sparse kernel based convolution for deep {CNNs},'' in \emph{2019
  57th Annual Allerton Conference on Communication, Control, and Computing
  (Allerton)}.\hskip 1em plus 0.5em minus 0.4em\relax IEEE, 2019, pp. 100--107.

\bibitem{horowitz20141}
M.~Horowitz, ``1.1 {Computing's} energy problem (and what we can do about
  it),'' in \emph{2014 IEEE International Solid-State Circuits Conference
  Digest of Technical Papers (ISSCC)}.\hskip 1em plus 0.5em minus 0.4em\relax
  IEEE, 2014, pp. 10--14.

\bibitem{shen_2017}
Y.~{Shen}, M.~{Ferdman}, and P.~{Milder}, ``Escher: A cnn accelerator with
  flexible buffering to minimize off-chip transfer,'' in \emph{2017 IEEE 25th
  Annual International Symposium on Field-Programmable Custom Computing
  Machines (FCCM)}, vol.~1, no.~1, 2017, pp. 93--100.

\bibitem{chen2015crossbar}
B.~{Chen}, F.~{Cai}, J.~{Zhou}, W.~{Ma}, P.~{Sheridan}, and W.~D. {Lu},
  ``Efficient in-memory computing architecture based on crossbar arrays,'' in
  \emph{2015 IEEE International Electron Devices Meeting (IEDM)}, vol.~1,
  no.~1, 2015, pp. 1--4.

\bibitem{loihi2018}
M.~{Davies}, N.~{Srinivasa}, T.~H. {Lin}, G.~{Chinya}, Y.~{Cao}, S.~H.
  {Choday}, G.~{Dimou}, P.~{Joshi}, N.~{Imam}, S.~{Jain}, Y.~{Liao}, C.~K.
  {Lin}, A.~{Lines}, R.~{Liu}, D.~{Mathaikutty}, S.~{McCoy}, A.~{Paul},
  J.~{Tse}, G.~{Venkataramanan}, Y.~H. {Weng}, A.~{Wild}, Y.~{Yang}, and
  H.~{Wang}, ``Loihi: A neuromorphic manycore processor with on-chip
  learning,'' \emph{IEEE Micro}, vol.~38, no.~1, pp. 82--99, 2018.

\bibitem{neuromorphicdataset2019}
S.~Miao, G.~Chen, X.~Ning, Y.~Zi, K.~Ren, Z.~Bing, and A.~Knoll, ``Neuromorphic
  vision datasets for pedestrian detection, action recognition, and fall
  detection,'' \emph{Frontiers in Neurorobotics}, vol.~13, p.~38, 2019.

\end{thebibliography}

\end{document}